# Generating Motion Patterns Using Evolutionary Computation in Digital Soccer

Masoud Amoozgar, Daniel Khashabi, Milad Heydarian, Mohammad Nokhbeh, Saeed Bagheri Shouraki

*Abstract*— Dribbling an opponent player in digital soccer environment is an important practical problem in motion planning. It has special complexities, which can be generalized to the most important problems in other similar multi-agent systems. In this paper, we propose a hybrid computational geometry and evolutionary computation approach for generating motion trajectories to avoid a mobile obstacle. In this case, an opponent agent is not only an obstacle but also one who tries to harden the dribbling procedure. One characteristic of this approach is reducing the process cost of online stage by transferring it to an offline stage, which causes increment in agents' performance. This approach breaks the problem into two offline and online stages. During the offline stage the goal is to find the desired trajectory using evolutionary computation and saving it as a trajectory plan. A trajectory plan consists of nodes, which approximate information of each trajectory plan. In the online stage a linear interpolation along with Delaunay Triangulation in a xy-plan is applied to the trajectory plan to retrieve the desired action.

## I. Introduction

THE research in Artificial Intelligence and its applications in Multi-agent Systems is one of the common fields of study in the recent years. The contribution of machine learning to learn and imitate complex patterns is therefore needed to make intelligent decisions based on sampled data. Some intelligent systems attempt to eliminate the need for human-intuition in data analysis, while others adopt a collaborative approach between human-intuition and machine. Human-intuition cannot, however, be entirely eliminated, since the system's designer must specify how the data is going to be represented and what mechanisms will be used to search for a characterization of the data. One other reason for collaborative human-machine approach is due to the complexity and difficulty of a pure machine analysis. In this study we are going to use a collaborative human-machine approach for designing a motion trajectory. Regarding to the fact that the way a mobile agent treating is somehow unpredictable before the online stage, human's intuition can be so effective for generating an initial dribbling behavior.

In this paper as a complex problem, analyzing the avoidance of a mobile obstacle is focused on. Obviously it is comparable with traditional obstacle avoidance problem. The manner of the obstacle in which it behaves has enormous effects on how the algorithm must be designed. The most important concern in this paper is the case in which obstacle agent tries to harden the motion task. The most glorious example of this problem is encountered in a digital soccer environment, where a ball-holder agent wants to dribble (avoid) a (several) defender agent(s), simultaneously reach a specific position or dribble toward specific direction.

In this paper, we propose a new approach for generating motion trajectories to avoid a mobile obstacle, which is hybrid of computational geometry and evolutionary computation. The offline stage is when the agent is out of competition; hence there is no strong limitation on competition time during this step. Therefore, the agent tries to explore better behaviors. In the online stage, an agent must exploit what it has learnt before. In fact, it is the time when the agent tries to act in the desired manner, so the online stage is limited to the whole game time. Transferring the process of calculation of the desired motion to the offline stage results decreasing of the computational cost of the online stage. Consequently, it causes increment of agent's performance. In the offline stage the goal is to find the desired trajectory plan using evolutionary computation. A trajectory plan consists of nodes, which approximate the information of each trajectory plan. In the online stage a linear interpolation along with Delaunay Triangulation in a xy-plan is applied to the trajectory plan to get the desired action.

In the rest of this paper, we first present some issues on dribbling a mobile agent. Then the offline stage and the online stage parts of the method are discussed. At the end of the paper, several results of the proposed method are presented. Dribble Editor (DEdit in brief) is used regarding generation of the motion trajectories. This new application enabled us designing the motion patterns and generating upcoming results. More pictures are depicted in (). The latest version of software and its source codes (in Java) are available under the following link:

`http://cs.sharif.edu/~amoozgar/focus/dedit`

## II. Dribbling a Mobile Agent

Dribbling is a sequence of several decisions, which determine the agent's position on a plan. In some cases the environmental position of an obstacle changes dynamically and a considered algorithm is supposed to be adaptive. But the way that obstacles behave is undeniable. In fact, if the rate of intelligence for an obstacle agent gets low, it won't be able to grab the ball. Our problem is that the ball-holder agent must avoid ball from the defender (opponent) agent. So it is required to keep the relative distance a little more than kickable radius. An obstacle agent could be more

Masood Amoozgar and Milad Heydarian are with Computer Science department, Sharif University of Technology. Email: {masood.moozgar,heydariaan}@cs.sharif.ir. Daniel Khashabi and Mohammad Nokhbeh are with Electrical Engineering Department, Amirkabir University of Technology. Email: {d.khashabi, nokhbeh100}@gmail.com. Saeed Bagheri Shouraki is associate professor of Sharif University of Technology, Electrical Engineering department. Email: bagheri-s@sharif.edu

intelligent and learn the way the dribbler is acting. So the actions of the dribbler agent are required to be adjusted greedier.

## III. OFFLINE STAGE

### A. Trajectory plan Generation

For generating a dribble trajectory we need to make a structure for defining a dribbling pattern. In fact, this structure is a function, which gives the required data of each position in the soccer field. Then the optimization procedure will be applied to this structure. In defining such a structure, we have used Delaunay Triangulation and Linear Interpolation, which are going to be described in the next parts.

*1) Delaunay Triangulation:* Delaunay triangulation is one of the most promising methods, which are used in order to triangulate a plate region, based on a given point-set. Delaunay Triangulation for a point-set $P$ is a triangulation $DT(P)$ so that no point in $P$ is inside the circumscribed circle of any triangle in $DT(P)$. For every set of points with the number greater than 3, there is a unique Delaunay Triangulation. In our data extraction method, vertices of triangles are agent's relative positions to the obstacle agent. For each vertex (position in the xy-plan), the information of the dribbler agent is already generated, e.g. direction of motion, direction of ball and acceleration of motion.

*2) Linear Interpolation:* For finding the information of a specific point $P_s$ inside of triangle $T_n$ (Fig. 1) we use simple linear interpolation. Assume $P_a$, $P_b$ and $P_c$ as vertices of a triangle with corresponding information $I_a$, $I_b$ and $I_c$, respectively. A weighted average for obtaining the information of point $P_s$ is:

$$I_s = \frac{\frac{I_a}{|P_s,P_a|} + \frac{I_b}{|P_s,P_b|} + \frac{I_C}{|P_s,P_c|}}{\frac{1}{|P_s,P_a|} + \frac{1}{|P_s,P_b|} + \frac{1}{|P_s,P_c|}} \quad (1)$$

The formula (1) can be considered as a conventional weighted mean of three values where $|P_i, P_j|$ is distance between two points $P_i$ and $P_j$. This kind of function approximation using Delaunay triangulation has the following advantages [ HYPERLINK \l "HAk" 1 ]:

--*High approximation accuracy:* It is claimed that this method of approximation is more accurate than other methods.

--*Locally adjustable:* Even if an existing data modified, the triangle region where the data is not in, is not affected.

--*Simple and fast*: Regarding to time complexity of triangulation (O*(nlogn)* see 2]} ) and linear interpolation, it is possible to use it in a real-time environment.

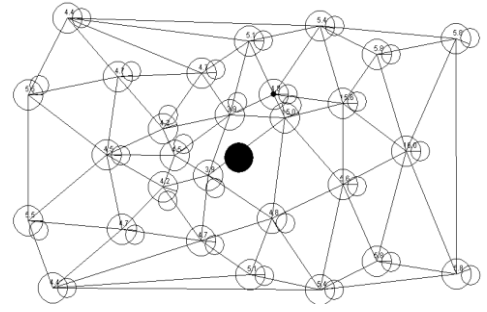

(a)

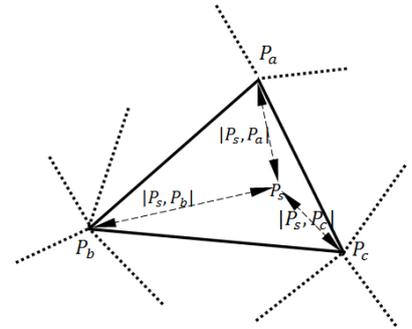

(b)

Fig. 1. (a) A Delaunay Triangulation applied on a point-set. White players are assumed to be dribbler agent and the black player is an obstacle. (b) Linear Interpolation applied in a triangle for retrieving information of a specific inner point e.g. $P_s$.

### B. Designing trajectory plans

Human intuition plays the prominent role in design of trajectory plan (See Fig. 4). In generating each pattern, the following attributes are considered:

--Position of an agent relative to the opponent agent
--Direction of agent's movement.
--Direction of ball.
--Acceleration of agent in the specified direction.

The position of the ball depends on its relative direction. After inserting each new node to a trajectory plan, DEdit triangulates the plan. It is also possible to edit specifications of the predefined nodes.

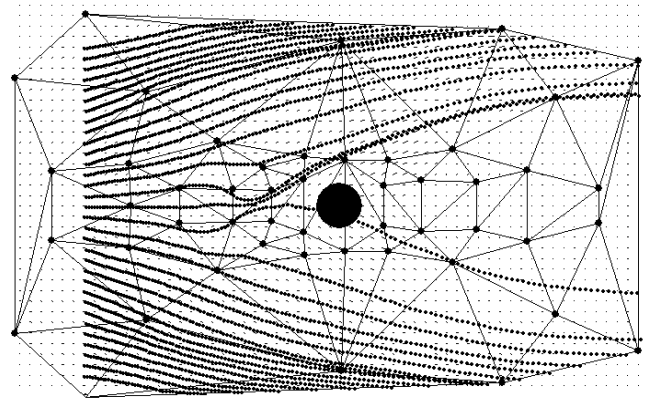

Fig. 2. Primary trajectory plan, which is manually designed in DEdit and simulated trajectories are generated consequently. Each simulation starts from x = -12 with initial velocity of 4 m/s.

For testing a trajectory plan a simulation engine is provided so that agent's movement is simulated starting from the initial position and velocity to a determined destination. An example of human-generated trajectory is presented in (Fig.2).

*C. Optimizing strategy of trajectory plans*

In order to apply Genetic Algorithm on trajectory plans, it is required to map the attributes of each plan to a string code. Each trajectory plan is introduced as an individual for the algorithm. Practically, there exist numerous nodes into a trajectory plan and each node includes 5 parameters: *x*, *y*, *acceleration*, *ball-direction* and *body-direction*. The parameters *x* and *y* are just stated in the purpose of distinguishing the points from others and are not considered in the evolution process.

The process of generating individuals based on the initial human-generated trajectory plan is done until reaching the size of population to the parameter *population-size*. This process generates the initial population. The next step of the algorithm will be repeated in the number of *generation-count*. In next parts, steps of evolution process are described in detail:

*[Step 1] Selecting several individuals as parents and copying them into mating pool:* Parent selection basically should better be a random process for diverging individuals in the search space. Moreover, better individuals (which are ranked by their fitness value) should gain more chance to be selected. Therefore, four algorithms are assumed:
  -- Roulette Wheel
  -- Rank
  -- Stochastic Universal Sampling
  -- Tournament
A parameter called *parent-selection-probability* controls the selection process.

*[Step 2] Applying crossover on individuals:* The crossover process consists of selecting two random parents and making two new individuals using parents. This process is repeated over and over until reaching the desired quantity. To control crossover process, a parameter *crossover-probability* is applied.

When two random individuals $P_i(g^{t-1})$ and $P_j(g^{t-1})$ are chosen from the parent group of previous generation $P(g^{t-1})$, two offspring $P_i(g^{t-1})^{child}$ and $P_j(g^{t-1})^{child}$ are generated by the crossover. This is carried out using the weighted mean, so $k^{th}$ parameter of the offspring individual is:

$$P_i(g^{t-1})^{child}.param_k = \frac{w_i \times P_i(g^{t-1}).param_k + w_j \times P_j(g^{t-1}).param_k}{w_i + w_j} \quad (2)$$

$$P_j(g^{t-1})^{child}.param_k = \frac{w_j \times P_i(g^{t-1}).param_k + w_i \times P_j(g^{t-1}).param_k}{w_i + w_j} \quad (3)$$

The parameters $w_1$ and $w_2$ are two random numbers between *0.8* and *1.2*.

*[Step 3] Mutating on population and creating new individuals and adding them into population:* Mutation process is defined by adding a Gaussian Random Number (*GRN*) to all parameters of each individual. These small random numbers are limited to *1%* of maximum value of the parameters so that for *body-direction* and *ball-direction*, the maximum value is *180°* and the corresponding *GNR* is a number between *-1.8* and *1.8*. In order to mutate individuals, an *initial mutation coefficient* is applied to create initial generation of individuals by mutation. Furthermore, the small random number multiplies *mutation-coefficient* to differentiate the primary mutation (which was applied on first individual to generate first population) and the next mutations. If $P_i(g^{t-1})^{mutated}$ is a mutated individual, the value of $k^{th}$ parameter of individual under mutation will be:

(4)
$$P_i(g^{t-1})^{mutated}.param_k = P_i(g^{t-1}).param_k + \left( \frac{\max\left[P(g^{t-1}).param_k\right] \times GRN \times mutationCoefficient}{100} \right)$$

*[Step 4] Selecting individuals from the population and transferring them to the next generation:* Selection of the individuals is a deterministic process i.e. transferring best *generation-number* of individuals to next generation. This is done after evaluating individuals based on a fitness function. Hence, this process results in maximizing fitness value of best individual, which leads to optimize the trajectory plan.

In the next part, structure of proposed fitness function is discussed.

*D. Fitness function*

Design of a fitness function, leads to determine the optimum behavior of the dribbler agent in each position of the plan. It is supposed that each node has a unique position in all trajectory plans. So the fitness function of a trajectory plan *tp* is the sum of the value of fitness function in included nodes.

$$planFitness(tp) = \sum_{i=1}^{\#nodes} fitnessFunction(tp.node_i) \quad (5)$$

And the value of fitness function for each node in the trajectory plan is obtained as following:

$$fitnessFunction(node) = \exp(desiredFunction(node)/\rho^2) \quad (6)$$

$$desiredFunction(node) = \alpha(node.bodyDir)^2 - (node.bodyDir - node.bodyRlObs)^2 - \beta(node.ballDir)^2 - (node.ballRlObs - \pi)^2 \quad (7)$$

$$\alpha = \alpha_{user}\left(0.1 + 50\exp(-2dist)\right)$$
$$\beta = \beta_{user}\left(0.1 + 50\exp(-2dist)\right) \quad (8)$$

Where in (7) the variables *bodyDir* and *ballDir* are body direction of agent and ball direction, respectively. *ballRlObs* is the angle between ball and the obstacle agent. *bodyRlObs* is the angle between dribbler agent and obstacle. Two factors are considered while designing the fitness function. The first is straightness and the second is safety of agent's motion. These factors are against each other. This is similar to what happens in the real soccer match. Direct movement is the fastest and low cost course and increases the excellence of ball control, whereas moving spirally increases safety of the motion and consequently leads to wasting energy and time. Therefore, balancing effectiveness of these factors in the fitness function will result the desired motion.

In order to adjust values of *bodyDir* and *ballDir* to desired ones, two factors are included for each of variables. First factor maximizes each variable in the function and the other one minimizes it. Rate of effectiveness for each of factors, depends on the corresponding coefficients ($\alpha$ and $\beta$). It is worth to mention that the user gives a certain value to $\alpha$ and $\beta$. Increment of $\alpha$ encourages the agent to move smoothly, which results in wasting less stamina and consequently, increment of $\beta$ encourages the agent to keep ball in front of its body, which results in maximizing ball control. Decrement of $\alpha$ and $\beta$ makes agent more sensitive to the opponent. For achieving a more similar behavior to the real agent, parameters $\alpha$ and $\beta$ have to be dependent from distance between agent and obstacle (Consider parameter *dist* in (8)).

## IV. ONLINE STAGE

### A. Using calculated values

In the online stage, dribbling action is contributed based on generated pattern. Depending on the position, in each cycle a trajectory plan determines the agent's acceleration, body-direction and ball-direction using linear interpolation. Reasons for proposing Linear Interpolation as the algorithm of calculating values of agent's motion is discussed in the rest. One can use other interpolating methods especially nonlinear approximations (e.g. Neural Networks or Fuzzy) depending on desired complexity.

Trajectory plans are obstacle-centric, which means position and direction of any trajectory plan in the digital soccer field, depends on obstacle (

Fig. 3). Therefore, in order to acquire an algorithm for choosing the best trajectory plan for agent's motion, Delaunay Triangulation is used for the second time. But in this layer of processing, nodes are assumed to be obstacles. In fact the trajectory plan in each position of the field is generated based on what is already explained. This is due to the fact that in different positions of the field, different manners of dribbling is desirable. Concluding, there are regions on the field so that safe action is needed, while in another region of the field low-cost and rapid action might be desired.

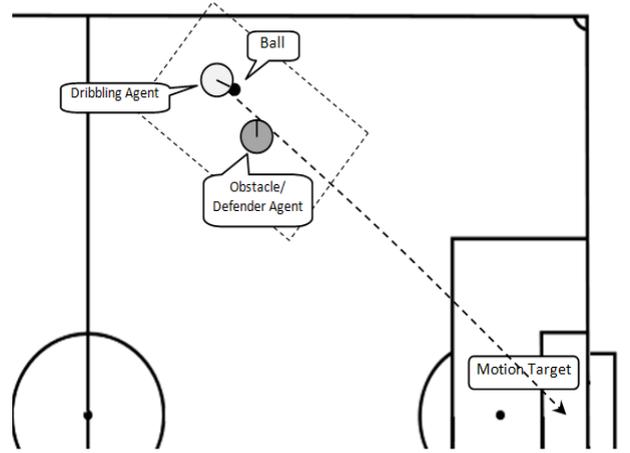

Fig. 3. Using a trajectory plan to generate the optimum behavior of dribbler agent. It depends on direction of obstacle relative to opponent's goal.

By encountering any obstacle, desired trajectory plan will be retrieved by applying Linear Interpolation on Delaunay mesh of soccer field based on saved trajectory plans, which is generated in the offline stage for various obstacle positions.

## V. RESULTS & FUTURE WORKS

In this study a method for designing motion trajectories is presented. The proposed method is a Genetic Algorithm, which is applied on solutions that are generated using Delaunay Triangulation. Though there is no strong proof on optimality of this trajectory generation, the results depicted in previous parts show that by adjusting fitness function and the use of offline dribble data set a trustable solution is achievable.

Though dribbling is a sequential action, the proposed method is free from sequential analysis. It is due to the fact that defining final results for dribbling is so complicated. So it would be difficult to apply methods like Reinforcement Learning to solve these kinds of problems. Results of generating desired patterns are shown in (Fig. 5), (Fig. 6) and (Fig. 7)

TABLE I.
PARAMETERS OF GENETIC ALGORITHM

| Parent selection algorithm | Crossover probability | Mutation coefficient | Parent selection coefficient | Generation count |
|---|---|---|---|---|
| *Roulette wheel* | 0.8 | 4 | 0.6 | 1000 |
| Max acceleration | Range of *bodyDir* | Range of *ballDir* | $\alpha_{user}$ | $\beta_{user}$ |
| 3 [m/s$^2$] | $[-\frac{\pi}{2}, \frac{\pi}{2}]$ | $[-\frac{\pi}{2}, \frac{\pi}{2}]$ | 0.66 | 0.33 |

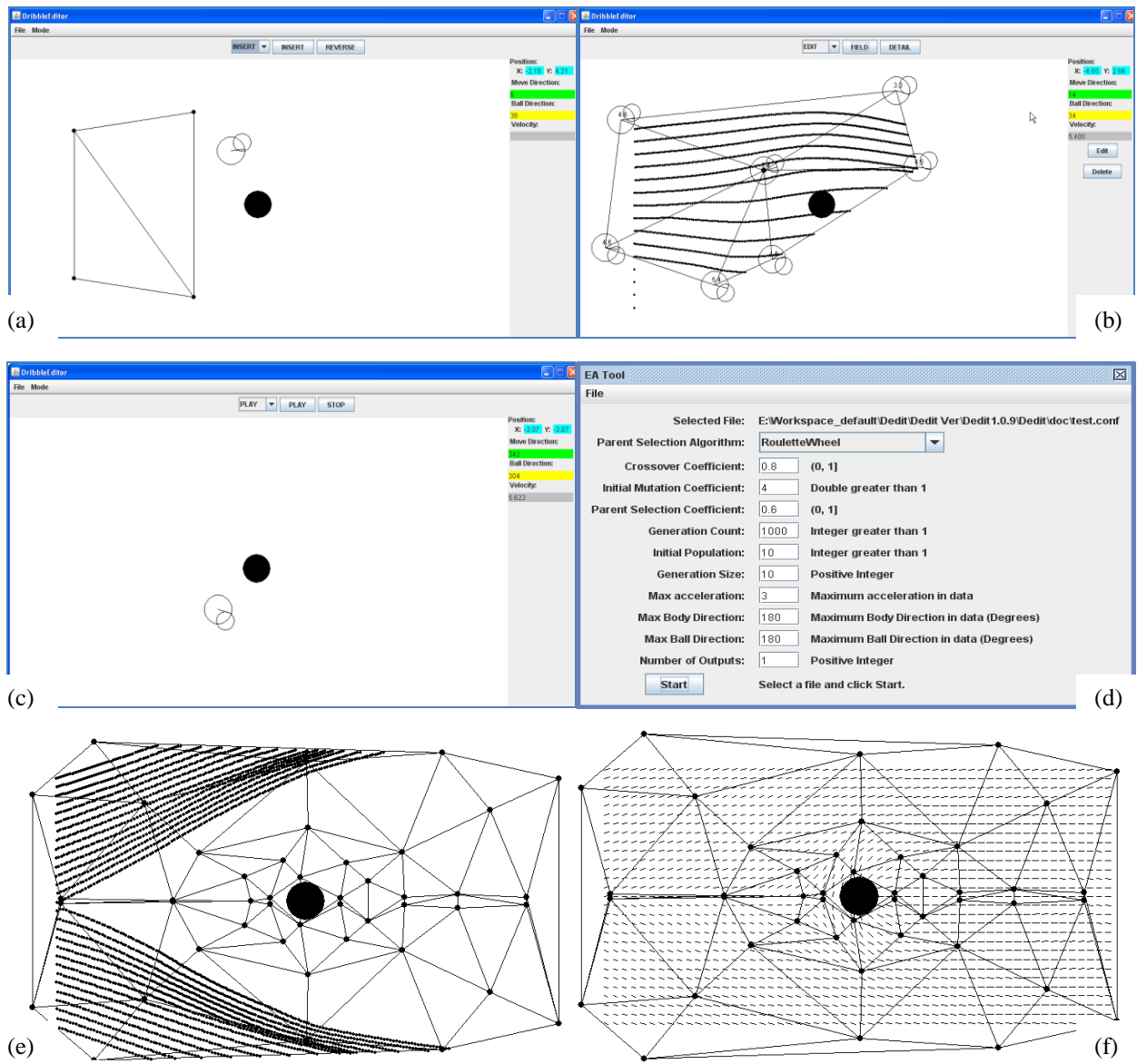

Fig. 4. (a) A screenshot of software used for generating motion patterns. INSERT mode (b) EDIT mode (c) PLAY mode. (d) Adjusting GA parameters. This application is designed to create, edit and test the trajectory plans. Each trajectory plan contains nodes with several independent variables: position (x, y), acceleration, body-direction and ball-direction (e) Resulted motion trajectories when agent started from different initial positions in the field (f) Direction and acceleration in each position of the field. It sounds similar to an electric field!

Getting more optimal results is what should be considered for the future works. In the online stage, the adaptive behavior is an important issue, where the dribbler agent changes its behavior based on its failures and successes. In general, the real world problems are dynamic and basis of action selection is to deceive the opponent. The implementing algorithm is required to be probabilistic and dependent to the reactions of the opponents. In advance steps, as the defense goes along legislating attacker's behavior, the reactions must be changed. So we are encountered with a dynamic action.

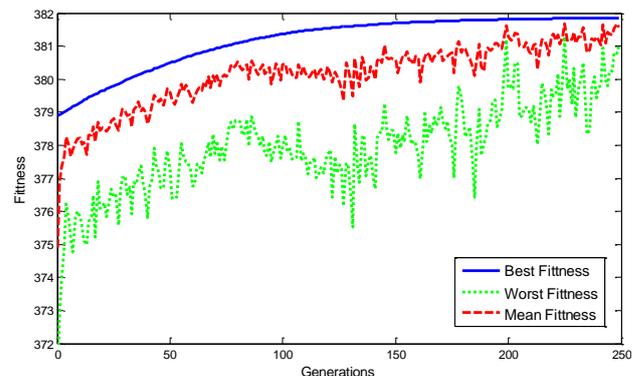

Fig. 5. Optimizing trajectory plans using EA. Best, mean and worst of fitness values in every generation during optimizing with parameters in

TABLE I. Values are averages of 8 times running.

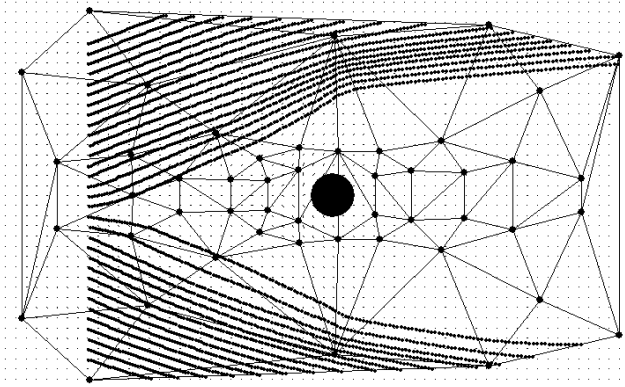

Fig. 6. Trajectory-plan and optimal trajectories after optimizing trajectory plan in Fig. 2, optimizing parameters came in TABLE I.

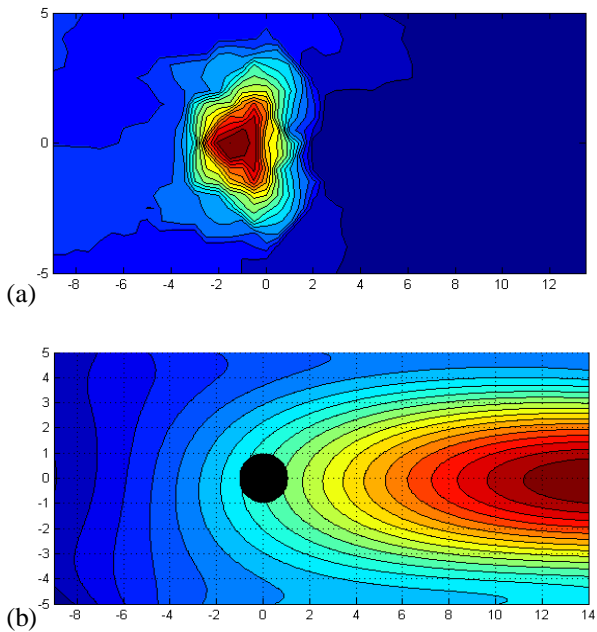

Fig. 7. (a) This figure shows distribution of acceleration in xy-plane. This picture is inferred from Fig.5 which is result of Genetic algorithm on motion field. In our implementation acceleration of each node is proportional to changes in velocity direction. As depicted acceleration near opponent has highest value. (b) Shows velocity distribution through the dribble field for a common initial speed for all initial position (x~-9). As depicted speed behind opponent player is increasing.

ACKNOWLEDGMENT

The authors would like to thank Mojtaba Khalidji for his illuminating comments and discussions. They also acknowledge Pooya Abedi, Iman Esmaili and Seyed Hossein Khasteh for their reviews and helpful comments.